\PassOptionsToPackage{table}{xcolor}
\documentclass[pdflatex,sn-nature]{sn-jnl}

\newcommand{\cfgTitle}{Hyper-Dimensional Fingerprints as Molecular Representations}
\newcommand{\cfgShortTitle}{Hyper-Dimensional Fingerprints}
\newcommand{\cfgKeywords}{Molecular Property Prediction, Molecular Fingerprints, Hyperdimensional Computing}

\newcommand{\cfgSnJnlAuthors}{%
    \author[1,2]{\fnm{Jonas} \sur{Teufel}\orcid{https://orcid.org/0000-0002-9228-9395}}\equalcont{These authors contributed equally to this work.}%
    \author[1,2]{\fnm{Luca} \sur{Torresi}\orcid{https://orcid.org/0000-0003-2205-6753}}\equalcont{These authors contributed equally to this work.}%
    \author[1,2]{\fnm{André} \sur{Eberhard}}%
    \author*[1,2]{\fnm{Pascal} \sur{Friederich}\orcid{https://orcid.org/0000-0003-4465-1465}}\email{pascal.friederich@kit.edu}%
    \affil[1]{\orgname{Karlsruhe Institute of Technology (KIT)}, \orgdiv{Institute of Nanotechnology (INT)}, \orgaddress{\street{Kaiserstraße 12}, \city{Karlsruhe}, \postcode{76131}, \country{Germany}}}%
    \affil[2]{\orgname{Karlsruhe Institute of Technology (KIT)}, \orgdiv{Institute of Anthropomatics and Robotics (IAR)}, \orgaddress{\street{Kaiserstraße 12}, \city{Karlsruhe}, \postcode{76131}, \country{Germany}}}%
}

\newcommand{\cfgDataAvailability}{All property-prediction datasets used in this study are publicly available through the \texttt{chem-mat-database} Python package \cite{chemmatdata}. Specific dataset identifiers are provided in the Supplementary Materials. The stereochemistry benchmarks of Supplementary Section~\siSecStereo{} are available from their original publications~\cite{adamsLearning3DRepresentations2022,baimachevaEvaluationChiralityDescriptors2025}.}
\newcommand{\cfgCodeAvailability}{The hyperdimensional fingerprint method is available as a Python package at \url{https://doi.org/10.5281/zenodo.19373620} under the MIT license. The complete code and pinned computational environment used to run the experiments and generate the results presented in this study are provided as a Code Ocean compute capsule under the MIT license, which reproduces representative results with a single ``Reproducible Run''; larger-scale runs are documented in the capsule. The capsule is made available to editors and reviewers during peer review through the journal's Code Ocean integration and will be published with a persistent DOI upon acceptance.}
\newcommand{\cfgAcknowledgements}{The authors acknowledge support from the state of Baden-Württemberg through bwHPC. Parts of this work were performed on the HoreKa supercomputer funded by the Ministry of Science, Research, and the Arts Baden-Württemberg and by the Federal Ministry of Education and Research. The authors thank the Gauss Centre for Supercomputing e.V. (GCS) for providing computing time on the Supercomputer JUPITER at the Jülich Supercomputing Centre via the John von Neumann Institute for Computing. JUPITER is supported by the EuroHPC JU and GCS through funding by the European Commission, the German Federal Ministry of Research, Technology and Space, and the Ministry of Culture and Science of the State of North Rhine-Westphalia. We acknowledge support by the Federal Ministry of Education and Research (BMBF) under Grant No. 01DM21002A (FLAIM). We acknowledge funding by the German Research Foundation (Deutsche Forschungsgemeinschaft, DFG) within Priority Programme SPP 2363. This work was supported by funding from the pilot program Core-Informatics of the Helmholtz Association (HGF).}
\newcommand{\cfgAuthorContributions}{J.T. designed and implemented the experimental pipeline and final software framework, performed the experimental evaluation, and wrote the initial manuscript draft. L.T. conceived the project, developed the core algorithm, and co-wrote the manuscript. A.E. contributed to the initial implementation and code optimization. P.F. supervised the project, secured funding, and edited the manuscript. J.T. and L.T. contributed equally to this work.}
\newcommand{\cfgCompetingInterests}{The authors declare no competing interests.}

\newcommand{\siSecSetupModelComparison}{S5.1}  
\newcommand{\siSecDimAblation}{S5.3}  
\newcommand{\siSecStereo}{S14}  
\newcommand{\siSecFingerprintComparison}{S5.2}  
\newcommand{\siSecDatasetsOpt}{S5}  
\newcommand{\siSecDescriptorAblation}{S9}  
\newcommand{\siSecGedEdit}{S11.1}  
\newcommand{\siSecGedAnalysis}{S11}  
\newcommand{\siSecBits}{S10}  
\newcommand{\siSecBoObjectives}{S13.2}  
\newcommand{\siSecBoProcedure}{S13.1}  
\newcommand{\siSecBoRuntime}{S13.3}  

\makeatletter
\@twosidefalse
\@mparswitchfalse
\makeatother

\usepackage[utf8]{inputenc}
\usepackage[T1]{fontenc}
\usepackage{etoolbox}

\usepackage{amsfonts}
\usepackage{amssymb}
\usepackage{amsmath}
\usepackage{nicefrac}
\usepackage{relsize}

\usepackage{booktabs}
\usepackage{graphicx}
\graphicspath{{figures/}}

\usepackage{xcolor}

\usepackage{microtype}
\usepackage{lipsum}

\hypersetup{colorlinks=true,linkcolor=black,citecolor=black,urlcolor=blue}

\usepackage{tikz}
\renewcommand{\orcidlogo}{\texorpdfstring{%
\raisebox{0.8ex}{\begin{tikzpicture}
\draw[green!50!black,fill=green!50!black] (0,0) circle (0.5ex);
\node[text=white,font=\sffamily\bfseries\fontsize{4}{4}\selectfont] at (0,0) {iD};
\end{tikzpicture}}%
}{}}

\definecolor{customblue}{RGB}{76,100,235}
\definecolor{customgreen}{RGB}{76,235,153}
\definecolor{custompurple}{RGB}{161,75,235}
\definecolor{customgray}{RGB}{128,128,128}
\definecolor{customlightblue}{RGB}{124,146,255}

\newcommand{\lightblue}{{\color{customlightblue}light blue}}
\newcommand{\blue}{{\color{customblue}blue}}
\newcommand{\green}{{\color{customgreen}green}}
\newcommand{\purple}{{\color{custompurple}purple}}
\newcommand{\gray}{{\color{customgray}gray}}

\newcommand{\qunit}[1]{{\footnotesize\color{darkgray}[#1]}}

\definecolor{trilowcolor}{RGB}{0,135,62}
\definecolor{trihighcolor}{RGB}{190,30,45}
\newcommand{\trilowsig}{\textcolor{trilowcolor}{\blacktriangledown}}
\newcommand{\trihighsig}{\textcolor{trihighcolor}{\blacktriangle}}
\newcommand{\trilow}{\textcolor{trilowcolor}{\triangledown}}
\newcommand{\trihigh}{\textcolor{trihighcolor}{\vartriangle}}

\title[\cfgShortTitle]{\cfgTitle}
\cfgSnJnlAuthors

\abstract{

Numerical molecular representations underpin virtual screening, property prediction, and materials discovery. Conventional fingerprints are efficient and deterministic but lose structural information through hash-based compression at low dimensionalities, whereas graph neural networks recover expressiveness through compute-intensive, task-specific training. Here, we introduce hyperdimensional fingerprints (HDF), which compose the chemical attributes of atoms and their multi-hop environments through algebraic operations on high-dimensional vectors, producing deterministic representations without any training. Across diverse property-prediction benchmarks, HDF ranks consistently among the best representations across datasets and models. Crucially, HDF encodes structure more efficiently: at low dimensionalities, its distances correlate with graph edit distance more strongly than Morgan fingerprints', yielding effective similarity metrics and competitive nearest-neighbor regression at just 64 components. In Bayesian optimization, HDF-based surrogates are more sample-efficient than Morgan fingerprints with a Tanimoto kernel for most of the tested objectives. These results establish HDF as compact, training-free representations for property prediction, similarity search, and sample-efficient optimization.
}


\begin{document}

\maketitle



\begin{figure}
    \centering
    \includegraphics[width=1\linewidth]{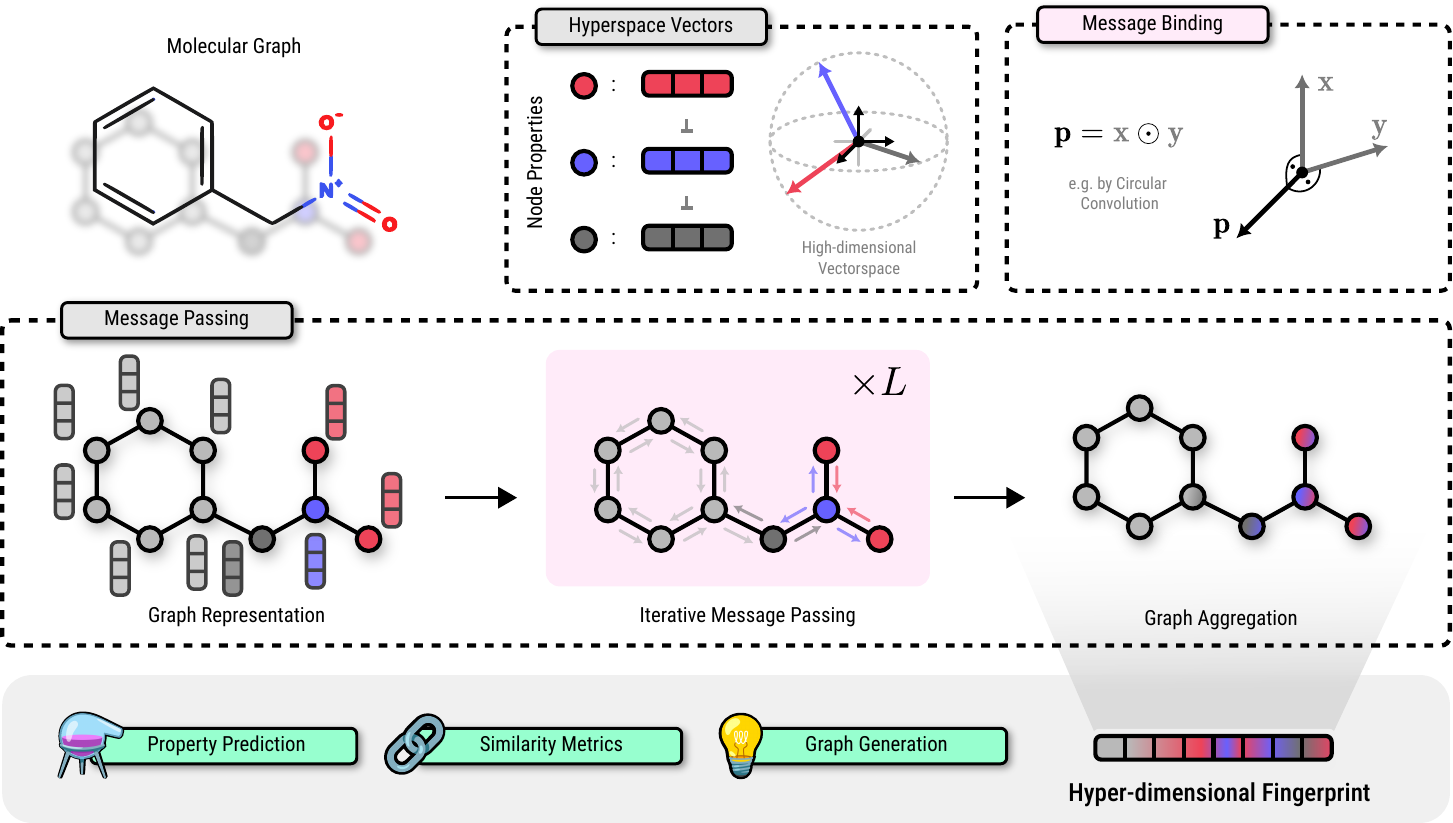}
    \caption{Overview of the hyperdimensional fingerprint representation. Atomic properties are mapped to high-dimensional vectors through predefined dictionaries and combined via circular convolution to form initial node embeddings. Iterative message passing propagates structural information across the molecular graph, and the final fingerprint is obtained by aggregating node embeddings across all iterations.}
    \label{fig:method-overview}
\end{figure}

Molecular fingerprints are fixed-size vector representations that encode the structural and chemical properties of molecules in a format suitable for computational analysis \cite{davidMolecularRepresentationsAIdriven2020,wighReviewMolecularRepresentation2022}. These representations are fundamental to modern cheminformatics, enabling property prediction, similarity-based virtual screening, and quantitative structure-activity relationship modeling. The ability to represent molecules of varying size as vectors of fixed dimensionality has made fingerprinting methods essential across applications ranging from solubility prediction to drug discovery.

Even as graph neural networks and large molecular foundation models have gained traction, molecular fingerprints continue to underpin a wide range of recent high-impact studies. Applications include antibiotic and antimicrobial discovery \cite{stokes2020antibiotic,wong2024structural,liu2023acinetobacter,scalia2025antibacterial,orsi2024ruthenium}, generative and de novo drug design \cite{moret2023clm,xie2025transpharmer}, data-driven catalyst and reaction optimization \cite{li2024photocatalyst,kingsmith2024reactome,zahrt2022electrochem,gotz2023htsynthesis,zhang2025rop}, inverse design of functional materials \cite{lyu2026perovskites}, and small-molecule property prediction in chemical biology \cite{ambadithody2024condensates}; recent surveys of medicinal chemistry, preclinical drug discovery, and natural-product drug discovery likewise treat fingerprint-based representations as a foundational tool \cite{mullowney2023natural,racz2025medchem,catacutan2024preclinical}. For example, Stokes et al.\ trained a deep neural network on Morgan fingerprints to identify halicin, a structurally distinct broad-spectrum antibiotic \cite{stokes2020antibiotic}, a strategy since extended to additional antibiotic classes \cite{wong2024structural,liu2023acinetobacter}. In a related domain, Li et al.\ used Morgan-fingerprint-based Tanimoto kernels within a closed-loop Bayesian optimization to identify metallophotocatalyst formulations competitive with iridium-based systems while exploring just 2.4\,\% of the candidate formulation space \cite{li2024photocatalyst}.

Traditional fingerprinting approaches, including substructure key-based fingerprints, path-based fingerprints, atom pair descriptors \cite{carhartAtomPairsMolecular1985}, and circular fingerprints such as Morgan fingerprints (also known as extended-connectivity fingerprints) \cite{rogersExtendedConnectivityFingerprints2010,morgan}, offer practical advantages including computational efficiency, deterministic computation, and interpretability. However, these methods rely on hash-based folding of substructural features into fixed-length bit vectors, which introduces information loss through bit collisions and limits representational capacity.

While learned molecular representations from graph neural networks \cite{gilmerNeuralMessagePassing2017,khemaniReviewGraphNeural2024} can achieve higher expressivity, they require training on task-specific data, demand substantial computational resources, and may not generalize well to domains outside their training distribution. This motivates the development of molecular representations that combine the efficiency and generality of traditional fingerprints with improved structural representation.

Hyperdimensional computing (HDC) \cite{kanervaHyperdimensionalComputingIntroduction2009,kleyko2022survey} uses the quasi-orthogonality of random vectors in high-dimensional spaces to build structured representations from simple algebraic operations, without any training. Recent work has begun to explore this framework for encoding graph-structured and molecular data. GraphHD \cite{nunes2022graphhd} and GrapHD \cite{poduval2022graphd} explored encoding graph topology into hypervector spaces, but both are restricted to immediate 1-hop neighborhood interactions, limiting the representation of multi-bond chemical environments. More recent methods formulate HDC graph encodings as message passing over multiple hops, such as VS-Graph~\cite{poursiami2025vs} for graph classification and HyperGraphX~\cite{cong2025hypergraphx} for node classification, and the graph vector function architecture (GVFA)~\cite{kahawala2026gvfa} provides a general framework for constructing graph representations without training, in which the learned transformations of message-passing neural networks are replaced by HDC operations. NysHD~\cite{zhao2025nyshd}, on the other hand, transfers established graph kernels to HDC through the Nystr\"{o}m method. In the molecular domain, MoleHD \cite{ma2022molehd} encodes molecules by tokenizing SMILES strings into hypervectors, processing them as linear sequences rather than graphs, while HDBind \cite{jones2024hdbind} applies HDC to molecular property prediction but relies on external feature extractors---such as conventional fingerprints or deep learning models---prior to hyperdimensional projection. Gobin et al.~\cite{gobin2024neuromorphic} adapted the 1-hop encoding of GrapHD to the atomic structures of materials to predict their bandgap. These approaches, along with further recent work \cite{verges2024molecular}, have established the viability of HDC for molecular and graph encoding tasks. However, the multi-hop HDC message-passing methods are designed for general graphs and have so far been evaluated on node and graph classification. To the best of our knowledge, it has not yet been investigated whether such representations can serve as molecular fingerprints that encode the chemical attributes of each atom and that can be used for property regression, similarity search and molecular optimization alike.

In this work, we introduce \textit{hyperdimensional fingerprints} (HDF), a molecular representation that integrates HDC with iterative message-passing frameworks and that can be seen as a formulation of the general principle of GVFA~\cite{kahawala2026gvfa} that is specialized for molecules (Figure~\ref{fig:method-overview}). HDF encodes atomic properties as randomly sampled hypervectors and propagates structural information through iterative binding and aggregation over the molecular graph, capturing multi-hop neighborhoods analogous to circular fingerprints of varying radii. Because the encoding rests on algebraic operations with well-understood mathematical properties rather than hash-based folding, HDF is entirely deterministic, requires no training, and avoids the bit-collision information loss that limits traditional fingerprints at reduced dimensionalities. Evaluating HDF across diverse molecular property-prediction benchmarks, we show that it matches or exceeds traditional fingerprinting methods while offering two distinguishing advantages: distances in the HDF embedding space correlate more strongly with graph edit distance, in particular at small embedding sizes, indicating that they reflect large structural differences between molecules more faithfully; and predictive performance is retained at substantially reduced embedding dimensions, with lower prediction errors than Morgan fingerprints of the same dimension at up to 256 components on the tested datasets. In Bayesian optimization of molecular properties, the compactness of HDF translates into improved sample efficiency: Gaussian process surrogates on HDF find near-optimal molecules in fewer evaluations than surrogates on Morgan fingerprints with the established Tanimoto kernel for six of nine objectives, and on these objectives, HDF with 16 to 64 dimensions, or 128 dimensions for QED, already matches the sample efficiency that Morgan fingerprints reach at their best embedding size. Together, these results establish HDF as a compact, training-free representation for property prediction, similarity search, and sample-efficient molecular optimization.


\section*{Results}

\begin{figure}[tb]
    \centering
    \includegraphics[width=\linewidth]{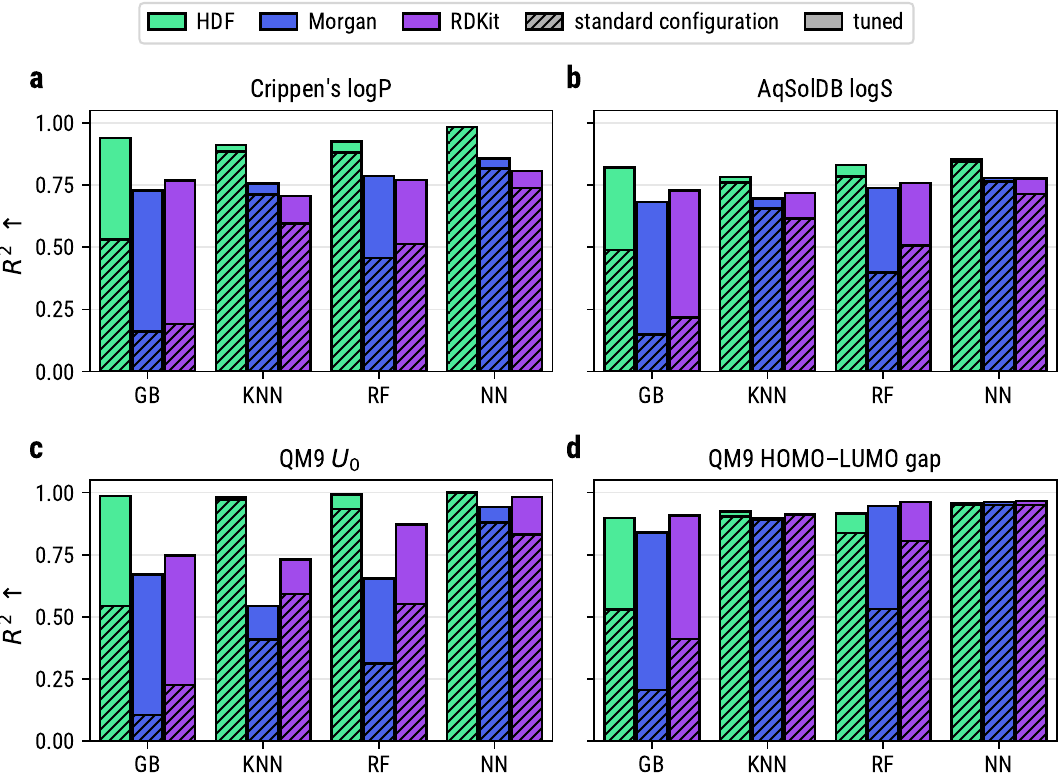}
    \caption{Model predictivity (R²) comparison of various simple machine learning models for 4 different molecular property regression tasks, (\textbf{a}) Crippen's logP, (\textbf{b}) AqSolDB logS, (\textbf{c}) the internal energy $U_0$ and (\textbf{d}) the HOMO-LUMO gap of QM9, when trained on Hyperdimensional Fingerprints (\green, ours), Morgan Fingerprints (\blue) and RDKit Fingerprints (\purple). The compared models are  Gradient Boosting (GB), K-nearest Neighbors (KNN), Random Forest (RF) and Multi-Layer Perceptrons (NN). Hatched bars show the result of the standard configuration of each representation and model, and the plain bars behind them the result after tuning representation and model together with the same budget for every combination, such that the visible plain part of a bar corresponds to the gain from tuning (see Supplementary Section~\siSecSetupModelComparison\ for details). Results are averaged over 10 independent repetitions.}
    \label{fig:ex01}
\end{figure}

In this section we present the results of various computational experiments and ablation studies meant to evaluate the performance of the proposed HDF representation in molecular property prediction tasks. We compare HDF against traditional Morgan fingerprints and other fingerprint methods implemented in the cheminformatics library RDKit \cite{landrum2006rdkit} to show their competitive performance across a variety of different property prediction tasks.


\subsection*{Property prediction with hyperdimensional fingerprints}
\label{sec:model_comparison}

We evaluate the effectiveness of HDF for molecular property prediction from two complementary perspectives: first, by examining performance across different machine learning paradigms to assess model-agnostic utility; and second, by comparing against established fingerprinting methods across a diverse set of chemical properties and datasets to assess generalization.

We first compare HDF against Morgan \cite{morgan} and RDKit \cite{landrum2006rdkit} fingerprints across four machine learning methods and four molecular property regression tasks (see Supplementary Materials Section~\siSecSetupModelComparison\ for details). Every representation is evaluated both in a standard configuration and after tuning it together with the model, using the same tuning budget of 30 configurations for all combinations of model and representation.

HDF substantially improves prediction performance across the majority of model-dataset pairs (Figure~\ref{fig:ex01}): It is the most accurate representation for 15 of the 16 pairs in the standard configuration and for 13 of the 16 pairs after tuning. Tuning improves all three representations, and it benefits Morgan and RDKit fingerprints more than HDF. Prediction of the internal energy $U_0$ exhibits the most substantial improvements, consistent with the fact that this quantum mechanical property depends primarily on local atomic environments and bond types---features that HDF represents through the element, the heavy-atom degree and the number of bonded hydrogens of each atom, which implicitly contain information on the bond types. On the HOMO-LUMO gap, by contrast, the tuned RDKit fingerprints are more accurate than HDF with three of the four models. We also observe that HDF performs particularly well with k-nearest neighbor methods, suggesting that its representation is particularly effective for molecular similarity tasks.

We next compare HDF against four established fingerprinting methods across a wider variety of property prediction datasets spanning diverse chemical domains and dataset sizes (see Supplementary Materials Section~\siSecFingerprintComparison{} for details on all methods and datasets).

When all representations use the same standard configuration, neural network models trained on HDF achieve the best result on 10 of the 17 property-dataset combinations, with a median rank of 1 (Table~\ref{tab:datasets}).
While no single method dominates across all tasks, HDF is the most consistent representation, which we consider its main benefit: In the geometric mean over all properties, its error deviates from the best result by 9\%, compared to 75\% to 222\% for the other fingerprints, and it lies within 10\% of the best result for 13 of the 17 properties. HDF is significantly more accurate than each of the four baselines (see Methods), and it again ranks first when every representation is tuned with the same budget (Supplementary Table~\siSecDatasetsOpt{}). The largest performance gaps between HDF and traditional fingerprints occur for extensive thermodynamic properties such as zero-point vibrational energy (ZPVE), enthalpy ($\Delta H$), and internal energy ($U_0$), where HDF achieves prediction errors up to two orders of magnitude smaller than hash-based fingerprints. These properties scale with molecular size and composition---information that HDF preserves through its atomic dictionaries encoding atom types and bond connectivity, but that hash-based methods lose through bit collisions when folding substructural features into fixed-length vectors. This advantage even persists when Morgan fingerprints are augmented with the graph size and the graph diameter, which HDF encodes as global descriptors (Supplementary Section~\siSecDescriptorAblation{}). This result highlights a key advantage of algebraic encoding over hashing: physically meaningful relationships between atomic environments are retained in the representation rather than compressed away.


\begin{table}
    \caption{Comparison of neural network models trained on different fingerprint representations across a variety of different properties and dataset sizes. All representations use the same standard configuration (2048 dimensions or bits, radius or depth 2) and the same neural network, and results are averaged over 10 independent repetitions with different data splits (Supplementary Section~\siSecFingerprintComparison{}). The best result in each row is shown in bold. $\trilowsig$/$\trihighsig$: significantly lower/higher MAE than HDF (two-sided Wilcoxon signed-rank test, paired by repetition, Holm-corrected within each target, $p < 0.05$); $\trilow$/$\trihigh$: lower/higher MAE than HDF without a significant difference. The first rows show the median rank of each method across all properties, the geometric mean of the relative deviation from the best result and the number of properties within 10\% of the best result. Additional information can be found in the Supplementary Materials.}
    \label{tab:datasets}
    \vspace{6pt}
    \centering
    \small
    \setlength{\tabcolsep}{2pt}
    \renewcommand{\arraystretch}{1.2}
\begin{tabular}{llrccccc}
\toprule
 &  &  & HDF & Morgan & RDKit & Torsion & Atom Pair \\
\midrule

\multicolumn{3}{l}{Median Rank} &
$1$ &
$3$ &
$5$ &
$4$ &
$2$ \\

\multicolumn{3}{l}{Geo.\ mean rel.\ deviation} &
$8.9\%$ &
$153.2\%$ &
$221.6\%$ &
$204.0\%$ &
$74.8\%$ \\

\multicolumn{3}{l}{Within 10\% of best} &
$13$ &
$7$ &
$4$ &
$2$ &
$8$ \\

\arrayrulecolor{gray}\midrule\arrayrulecolor{black}

Dataset & Quantity & Datapoints & HDF & Morgan & RDKit & Torsion & Atom Pair \\

\arrayrulecolor{gray}\midrule\arrayrulecolor{black}

HOPV15 &
PCE \qunit{\%} &
176 &
$\underset{\color{darkgray} \pm0.35}{ 1.788 }$ &
$\underset{\color{darkgray} \pm0.33}{ 1.447 ^{\trilowsig}}$ &
$\underset{\color{darkgray} \pm0.28}{ 1.482 ^{\trilowsig}}$ &
$\underset{\color{darkgray} \pm0.26}{ 1.552 ^{\trilowsig}}$ &
$\underset{\color{darkgray} \pm0.28}{\mathbf{ 1.421 }^{\trilowsig}}$ \\

 &
$V_\mathrm{oc}$ \qunit{V} &
176 &
$\underset{\color{darkgray} \pm0.019}{ 0.1135 }$ &
$\underset{\color{darkgray} \pm0.026}{ 0.0984 ^{\trilow}}$ &
$\underset{\color{darkgray} \pm0.024}{ 0.1163 ^{\trihigh}}$ &
$\underset{\color{darkgray} \pm0.027}{ 0.1124 ^{\trilow}}$ &
$\underset{\color{darkgray} \pm0.020}{\mathbf{ 0.0975 }^{\trilowsig}}$ \\

FreeSolv &
$\Delta G$ \qunit{kcal/mol} &
639 &
$\underset{\color{darkgray} \pm0.10}{\mathbf{ 0.824 }}$ &
$\underset{\color{darkgray} \pm0.20}{ 1.239 ^{\trihighsig}}$ &
$\underset{\color{darkgray} \pm0.12}{ 0.996 ^{\trihighsig}}$ &
$\underset{\color{darkgray} \pm0.13}{ 1.768 ^{\trihighsig}}$ &
$\underset{\color{darkgray} \pm0.10}{ 1.161 ^{\trihighsig}}$ \\

BACE &
IC$_{50}$ \qunit{pIC$_{50}$} &
1{,}513 &
$\underset{\color{darkgray} \pm0.05}{ 0.544 }$ &
$\underset{\color{darkgray} \pm0.05}{ 0.512 ^{\trilowsig}}$ &
$\underset{\color{darkgray} \pm0.05}{ 0.546 ^{\trihigh}}$ &
$\underset{\color{darkgray} \pm0.05}{ 0.522 ^{\trilow}}$ &
$\underset{\color{darkgray} \pm0.03}{\mathbf{ 0.505 }^{\trilow}}$ \\

Lipophilicity &
logD &
4{,}199 &
$\underset{\color{darkgray} \pm0.01}{\mathbf{ 0.490 }}$ &
$\underset{\color{darkgray} \pm0.02}{ 0.563 ^{\trihighsig}}$ &
$\underset{\color{darkgray} \pm0.02}{ 0.596 ^{\trihighsig}}$ &
$\underset{\color{darkgray} \pm0.02}{ 0.592 ^{\trihighsig}}$ &
$\underset{\color{darkgray} \pm0.02}{ 0.545 ^{\trihighsig}}$ \\

AqSolDB &
logS \qunit{log\,M} &
9{,}889 &
$\underset{\color{darkgray} \pm0.02}{\mathbf{ 0.606 }}$ &
$\underset{\color{darkgray} \pm0.03}{ 0.753 ^{\trihighsig}}$ &
$\underset{\color{darkgray} \pm0.03}{ 0.848 ^{\trihighsig}}$ &
$\underset{\color{darkgray} \pm0.03}{ 0.773 ^{\trihighsig}}$ &
$\underset{\color{darkgray} \pm0.01}{ 0.619 ^{\trihighsig}}$ \\

 &
ClogP &
9{,}889 &
$\underset{\color{darkgray} \pm0.01}{\mathbf{ 0.181 }}$ &
$\underset{\color{darkgray} \pm0.03}{ 0.550 ^{\trihighsig}}$ &
$\underset{\color{darkgray} \pm0.03}{ 0.662 ^{\trihighsig}}$ &
$\underset{\color{darkgray} \pm0.02}{ 0.520 ^{\trihighsig}}$ &
$\underset{\color{darkgray} \pm0.02}{ 0.308 ^{\trihighsig}}$ \\

COMPAS-3X &
$U_0$ \qunit{eV} &
39{,}482 &
$\underset{\color{darkgray} \pm0.17}{\mathbf{ 0.846 }}$ &
$\underset{\color{darkgray} \pm1.83}{ 97.089 ^{\trihighsig}}$ &
$\underset{\color{darkgray} \pm2.02}{ 134.333 ^{\trihighsig}}$ &
$\underset{\color{darkgray} \pm2.16}{ 89.114 ^{\trihighsig}}$ &
$\underset{\color{darkgray} \pm1.44}{ 77.891 ^{\trihighsig}}$ \\

 &
Gap \qunit{eV} &
39{,}482 &
$\underset{\color{darkgray} \pm0.003}{ 0.1595 }$ &
$\underset{\color{darkgray} \pm0.001}{\mathbf{ 0.0887 }^{\trilowsig}}$ &
$\underset{\color{darkgray} \pm0.003}{ 0.2270 ^{\trihighsig}}$ &
$\underset{\color{darkgray} \pm0.002}{ 0.1780 ^{\trihighsig}}$ &
$\underset{\color{darkgray} \pm0.005}{ 0.2212 ^{\trihighsig}}$ \\

 &
$\mu$ \qunit{D} &
39{,}482 &
$\underset{\color{darkgray} \pm0.003}{ 0.0478 }$ &
$\underset{\color{darkgray} \pm0.001}{\mathbf{ 0.0351 }^{\trilowsig}}$ &
$\underset{\color{darkgray} \pm0.001}{ 0.0542 ^{\trihighsig}}$ &
$\underset{\color{darkgray} \pm0.002}{ 0.0518 ^{\trihighsig}}$ &
$\underset{\color{darkgray} \pm0.002}{ 0.0524 ^{\trihighsig}}$ \\

QM9 &
$\mu$ \qunit{D} &
133{,}882 &
$\underset{\color{darkgray} \pm0.00}{\mathbf{ 0.558 }}$ &
$\underset{\color{darkgray} \pm0.01}{ 0.575 ^{\trihighsig}}$ &
$\underset{\color{darkgray} \pm0.01}{ 0.579 ^{\trihighsig}}$ &
$\underset{\color{darkgray} \pm0.01}{ 0.673 ^{\trihighsig}}$ &
$\underset{\color{darkgray} \pm0.01}{ 0.594 ^{\trihighsig}}$ \\

 &
$C_v$ \qunit{cal/(mol\,K)} &
133{,}882 &
$\underset{\color{darkgray} \pm0.00}{ 0.332 }$ &
$\underset{\color{darkgray} \pm0.01}{ 0.822 ^{\trihighsig}}$ &
$\underset{\color{darkgray} \pm0.01}{ 1.566 ^{\trihighsig}}$ &
$\underset{\color{darkgray} \pm0.01}{ 1.009 ^{\trihighsig}}$ &
$\underset{\color{darkgray} \pm0.00}{\mathbf{ 0.321 }^{\trilowsig}}$ \\

 &
ZPVE \qunit{mHa} &
133{,}882 &
$\underset{\color{darkgray} \pm0.02}{\mathbf{ 0.824 }}$ &
$\underset{\color{darkgray} \pm0.05}{ 4.957 ^{\trihighsig}}$ &
$\underset{\color{darkgray} \pm0.10}{ 8.418 ^{\trihighsig}}$ &
$\underset{\color{darkgray} \pm0.09}{ 6.065 ^{\trihighsig}}$ &
$\underset{\color{darkgray} \pm0.03}{ 1.565 ^{\trihighsig}}$ \\

 &
$\alpha$ \qunit{$a_0^3$} &
133{,}882 &
$\underset{\color{darkgray} \pm0.01}{\mathbf{ 0.625 }}$ &
$\underset{\color{darkgray} \pm0.03}{ 1.848 ^{\trihighsig}}$ &
$\underset{\color{darkgray} \pm0.04}{ 2.646 ^{\trihighsig}}$ &
$\underset{\color{darkgray} \pm0.02}{ 2.263 ^{\trihighsig}}$ &
$\underset{\color{darkgray} \pm0.01}{ 0.679 ^{\trihighsig}}$ \\

 &
$\Delta H$ \qunit{Ha} &
133{,}882 &
$\underset{\color{darkgray} \pm0.05}{\mathbf{ 0.550 }}$ &
$\underset{\color{darkgray} \pm0.21}{ 7.332 ^{\trihighsig}}$ &
$\underset{\color{darkgray} \pm0.15}{ 9.163 ^{\trihighsig}}$ &
$\underset{\color{darkgray} \pm0.11}{ 10.077 ^{\trihighsig}}$ &
$\underset{\color{darkgray} \pm0.05}{ 1.369 ^{\trihighsig}}$ \\

 &
$U_0$ \qunit{Ha} &
133{,}882 &
$\underset{\color{darkgray} \pm0.04}{\mathbf{ 0.518 }}$ &
$\underset{\color{darkgray} \pm0.18}{ 7.351 ^{\trihighsig}}$ &
$\underset{\color{darkgray} \pm0.13}{ 9.173 ^{\trihighsig}}$ &
$\underset{\color{darkgray} \pm0.13}{ 10.065 ^{\trihighsig}}$ &
$\underset{\color{darkgray} \pm0.06}{ 1.354 ^{\trihighsig}}$ \\

 &
Gap \qunit{Ha} &
133{,}882 &
$\underset{\color{darkgray} \pm0.0001}{ 0.00729 }$ &
$\underset{\color{darkgray} \pm0.0001}{ 0.00717 ^{\trilowsig}}$ &
$\underset{\color{darkgray} \pm0.0001}{ 0.00704 ^{\trilowsig}}$ &
$\underset{\color{darkgray} \pm0.0001}{ 0.00802 ^{\trihighsig}}$ &
$\underset{\color{darkgray} \pm0.0001}{\mathbf{ 0.00687 }^{\trilowsig}}$ \\

\bottomrule
\end{tabular}

\end{table}



\subsection*{Hyperdimensional fingerprints capture structural similarity}

\begin{figure}[t]
    \centering
    \includegraphics[width=\linewidth]{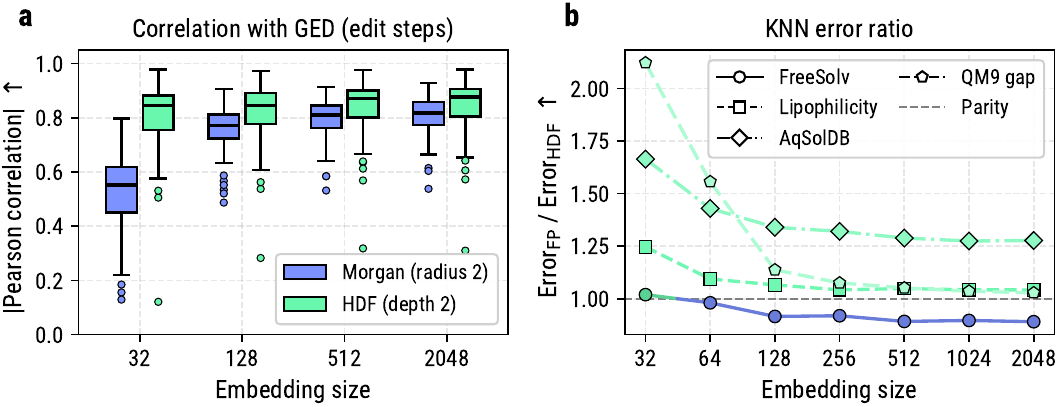}
    \caption{\textbf{\textsf{a}} Absolute Pearson correlation between fingerprint-based distances and graph edit distance (GED) across embedding sizes, comparing Morgan fingerprints with radius 2 (\blue) and hyperdimensional fingerprints with depth 2 (\green, ours). Box plots summarize the distribution over 100 seed molecules.  \textbf{\textsf{b}} Ratio of K-nearest neighbor prediction error for Morgan fingerprints to hyperdimensional fingerprints ($\mathrm{Error}_{\mathrm{FP}} / \mathrm{Error}_{\mathrm{HDF}}$) as a function of embedding dimensionality for four molecular property datasets. A dashed line at 1.0 marks parity; values above 1.0 indicate that hyperdimensional fingerprints (\green, ours) achieve lower prediction error than Morgan fingerprints. Both methods use a fixed radius/depth of 2 and $k = 5$ neighbors with the native distance of each representation (Tanimoto for Morgan fingerprints, cosine for hyperdimensional fingerprints), and the ratios are averaged over 10 repetitions.}
    \label{fig:ex_bio}
\end{figure}

The preceding experiments demonstrate that HDF achieves strong predictive performance across diverse molecular property tasks. A natural question is whether this performance arises from HDF capturing meaningful structural relationships between molecules. We hypothesize that distances in the HDF embedding space more accurately reflect molecular topology than traditional fingerprints, and that this property directly benefits distance-dependent prediction methods. We present two complementary experiments to support this hypothesis.

We first analyze the correlation between representation-space distances and graph edit distance (GED) \cite{sanfeliuDistanceMeasureAttributed1983}. Since exact GED computation is NP-hard for general molecular graphs, we use the number of edit steps as a proxy for the GED: Starting from randomly selected seed molecules, we iteratively apply chemically valid single edits (atom additions, bond removals and bond-order reductions) and sample molecules at one to three edit steps from each seed (see Supplementary Materials Section~\siSecGedEdit{} for the detailed procedure). We then compute the cosine distance for HDF vectors and the Tanimoto distance for Morgan fingerprints at each embedding dimensionality. We provide additional analyses of the correlation between representation distances and GED in Supplementary Section~\siSecGedAnalysis{}.

In this edit-based analysis, HDF achieves a higher Pearson correlation with the number of edit steps than Morgan fingerprints of the same dimension at all tested embedding sizes (Figure~\ref{fig:ex_bio}a). At 32 dimensions, HDF attains a median correlation of approximately 0.85, compared to approximately 0.55 for Morgan fingerprints. As embedding size increases, both representations improve, but HDF maintains its advantage: at 512 and 2048 dimensions, HDF achieves median correlations approaching 0.9, compared to approximately 0.8 for Morgan fingerprints. For the exact GED of random pairs of small molecules, HDF also correlates more strongly than Morgan fingerprints, in particular at small embedding sizes (Supplementary Section~\siSecGedAnalysis{}).

As a downstream consequence of this improved structural fidelity, we would expect HDF to benefit distance-based prediction methods on properties that vary smoothly with molecular topology. To probe this, we compare K-nearest neighbor regressors ($k = 5$) trained on HDF and Morgan fingerprints across embedding dimensionalities from 32 to 2048 and report the error ratio $\mathrm{MAE}_{\mathrm{Morgan}} / \mathrm{MAE}_{\mathrm{HDF}}$ (Figure~\ref{fig:ex_bio}b). For most datasets, HDF yields lower KNN errors than Morgan fingerprints, with the largest gains at low embedding sizes where the GED correlation gap is also widest. The effect is naturally dataset-dependent: not every molecular property varies smoothly with topological distance, and for properties driven by features that happen to be well-captured by Morgan's substructure hashing, the ratio can fall near or below parity. The overall trend nevertheless supports the interpretation that faithful distance structure is a meaningful source of HDF's advantage.


\subsection*{Hyperdimensional fingerprints remain accurate at low dimensions}

\begin{figure}[t]
    \centering
    \includegraphics[width=\linewidth]{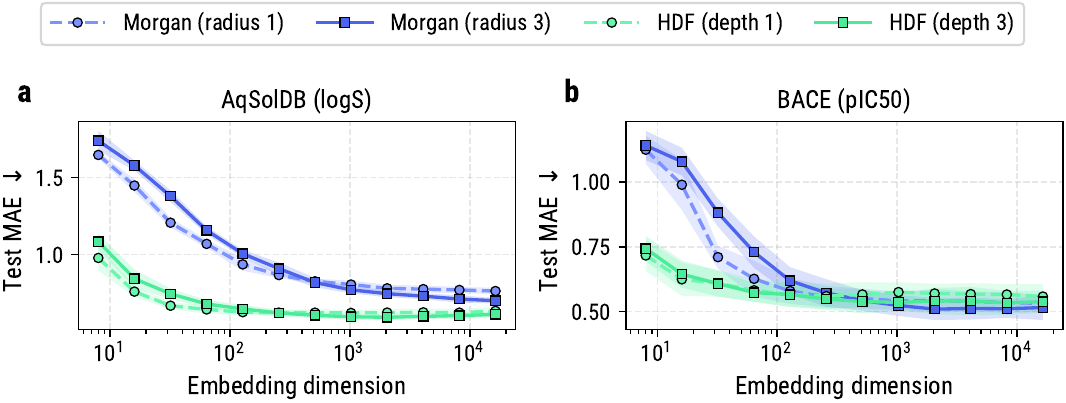}
    \caption{Prediction error (MAE) of neural network (NN) regressors trained on Morgan fingerprints (\blue) and hyperdimensional fingerprints (\green, ours) as a function of embedding dimensionality, on (\textbf{a}) AqSolDB (logS) and (\textbf{b}) BACE (pIC50). Each panel compares both representations at two message-passing depths (1 and 3 for HDF; matching radii for Morgan). Results show the mean and standard deviation over 10 independent repetitions.}
    \label{fig:ex_size}
\end{figure}

We investigate the relationship between embedding dimensionality and predictive performance by evaluating HDF and Morgan fingerprints with neural network regressors at dimensions ranging from 8 to 16,384 components (see Supplementary Materials Section~\siSecDimAblation\ for details).

At low dimensionalities (up to 256 components), HDF achieves lower prediction errors than Morgan fingerprints of the same dimension across the tested datasets (Figure~\ref{fig:ex_size}a,b). This advantage reflects the different ways the two representations handle limited dimensionality: Morgan fingerprints hash substructural features into a fixed-length bit vector and therefore incur collision-induced information loss that is most severe at small embedding sizes, whereas HDF distributes structural information continuously across the embedding through algebraic operations. As the dimensionality grows, the Morgan collision rate decreases and the gap narrows: on AqSolDB HDF remains more accurate at all dimensions, while on BACE Morgan slightly outperforms HDF at higher dimensions.  We compare HDF and Morgan fingerprints at the same number of dimensions, which determines the input size of the downstream models, although an HDF vector in 32-bit floating-point format requires 32 times the memory of a binary fingerprint of the same dimension. Since HDF with 2048 dimensions retains its predictive performance on the tested datasets when quantized to 8 bits per dimension and loses little at 4 bits, this overhead can be reduced to a factor of about 4 to 8 (Supplementary Section~\siSecBits{}).


\subsection*{Bayesian optimization with hyperdimensional fingerprints}

\begin{figure}[!tp]
    \centering
    \includegraphics[width=1.0\linewidth]{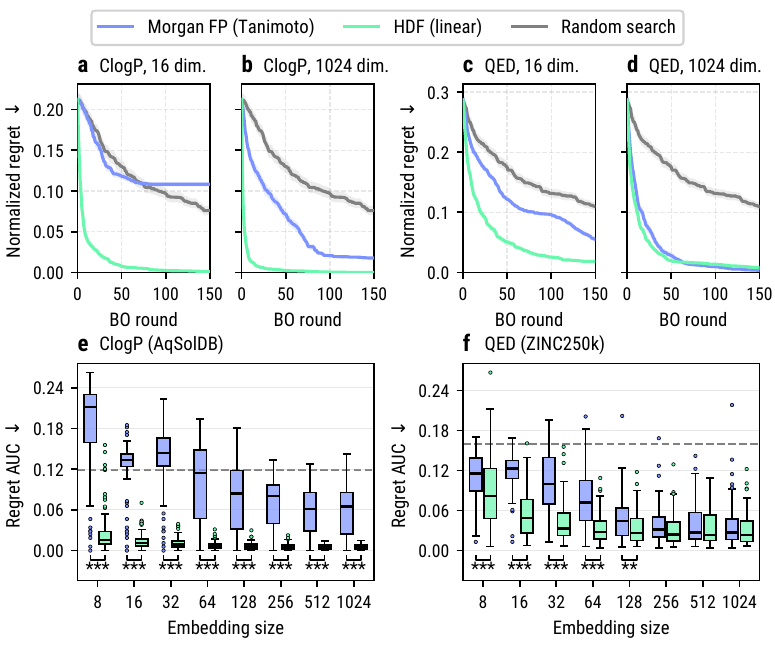}
    \caption{Bayesian optimization of molecular properties with different molecular representations and Gaussian process kernels. All results aggregate 100 independent optimization campaigns per representation and embedding size. \textsf{\textbf{a--d}} Normalized regret of the best molecule found so far over the optimization rounds (lower is better) for a target ClogP value of 15 on the molecules of AqSolDB at \textsf{\textbf{(a)}} 16 and \textsf{\textbf{(b)}} 1024 dimensions and for a target QED value of 0.2 on ZINC250k at \textsf{\textbf{(c)}} 16 and \textsf{\textbf{(d)}} 1024 dimensions. Curves show the mean for Morgan fingerprints with radius~2 and a Tanimoto kernel (\lightblue), hyperdimensional fingerprints with depth~2 and a linear kernel (\green, ours) and random search (\gray); shaded bands denote standard error. \textsf{\textbf{e,\,f}} Area under the regret curve (regret AUC, lower is better) across embedding sizes from 8 to 1024 for \textsf{\textbf{(e)}} ClogP and \textsf{\textbf{(f)}} QED. Box plots show the spread over campaigns, and the dashed line marks the median of random search. Brackets indicate statistical significance between HDF and Morgan fingerprints with the Tanimoto kernel (two-sided Wilcoxon signed-rank test, Holm-corrected, *\,$p<0.05$, **\,$p<0.01$, ***\,$p<0.001$).}
    \label{fig:ex_bo}
\end{figure}

The preceding experiments demonstrate two key properties of the HDF representation: (a) distances in the embedding space exhibit stronger correlation with structural similarity compared to traditional fingerprints, and (b) predictive performance of HDF representations stays high even at substantially reduced embedding dimensions. These characteristics are particularly relevant for Bayesian optimization, where low-dimensional representations are preferred due to the computational cost of Gaussian process inference, and where the surrogate model benefits from smooth mappings between representation space and target properties.

To assess whether these properties translate to improved sample efficiency, we design a Bayesian optimization benchmark that simulates the search for molecules with optimal property values in a fixed candidate library. Each optimization campaign starts from 20 randomly selected molecules and, in each of 150 rounds, fits a Gaussian process surrogate model to all observed values and evaluates the candidate with the highest log expected improvement~\cite{ament2023logei} among all remaining molecules of the library. Since the surrogate compares molecules exclusively through its kernel function, we pair each representation with a kernel that suits its structure: binary Morgan fingerprints with radius~2 use the Tanimoto kernel, the established choice for bit vectors~\cite{ralaivola2005graph,griffiths2023gauche}, whereas HDF uses a linear kernel. Random search, which selects candidates without any surrogate model, serves as the reference. We quantify performance by the normalized regret, the gap between the best value found so far and the best value of the library relative to the range of values in the library, and summarize each campaign by the area under its regret curve (regret AUC), such that lower values indicate that good molecules are found earlier. Figure~\ref{fig:ex_bo} shows the results for a target ClogP value of 15 on the molecules of AqSolDB and a target QED value of 0.2 on ZINC250k, with 100 campaigns per representation and embedding size between 8 and 1024 dimensions; seven further objectives, ranging from measured solubilities and melting points to DFT-computed electronic properties, as well as two further surrogates with a Mat\'{e}rn kernel, the principal components of 2048-bit Morgan fingerprints and HDF, are reported in Supplementary Section~\siSecBoObjectives{}, and the detailed procedure is described in Supplementary Section~\siSecBoProcedure{}.

For both objectives, HDF finds near-optimal molecules faster than Morgan fingerprints with the Tanimoto kernel, most clearly at small embedding sizes (Figure~\ref{fig:ex_bo}). For ClogP, this advantage holds at every embedding size (at 16 dimensions, all 100 HDF campaigns reach a normalized regret of at most 0.05, compared to 8 of 100 for Morgan fingerprints), whereas for QED, the difference becomes small from 256 dimensions on.

HDF is more sample-efficient than Morgan fingerprints with the Tanimoto kernel on six of the nine objectives, whereas Morgan fingerprints perform better on one of them and neither representation is consistently better on the remaining two (Supplementary Section~\siSecBoObjectives{}). On these six objectives, HDF with only 16 to 64 dimensions already matches the best sample efficiency that Morgan fingerprints reach at any embedding size (128 dimensions for QED). We attribute this to the dimensional efficiency of HDF: Folded fingerprints lose information through bit collisions at small embedding sizes, whereas HDF continues to encode the whole molecular graph. Since the cost of every optimization round grows with the embedding size, compact representations additionally reduce the cost of Bayesian optimization (on ZINC250k, a campaign takes a few minutes at 8 dimensions but two to three hours at 512 dimensions; Supplementary Section~\siSecBoRuntime{}).


\section*{Discussion}

The results presented above demonstrate that hyperdimensional fingerprints provide a compact, training-free molecular representation in which embedding distances reflect molecular topology more faithfully than those of Morgan fingerprints. This structural fidelity, quantified by a stronger correlation with graph edit distance, is the central property from which the main practical advantages of HDF follow.

First, the faithful preservation of topological similarity directly explains the strong performance of HDF with distance-dependent methods. K-nearest neighbor regressors, which rely entirely on pairwise distances in the embedding space, tend to benefit from HDF's distance structure compared to Morgan fingerprints at matched dimensionalities. This result suggests that HDF embeddings induce a metric space in which chemically related molecules are consistently placed closer together, a property that hash-based fingerprints do not guarantee.

Second, HDF maintains this structural fidelity at remarkably low dimensionalities. At up to 256 components, HDF achieves lower prediction errors than Morgan fingerprints of the same dimension, and distances retain most of their correlation with graph edit distance even at 32 dimensions, whereas the correlation of Morgan fingerprints drops considerably. This dimensional efficiency carries over to Bayesian optimization: On six of nine objectives, Gaussian process surrogates on HDF with 16 to 64 dimensions, or 128 dimensions for QED, match or improve on the median sample efficiency that Morgan fingerprints with a Tanimoto kernel reach at their best embedding size.

However, these advantages do not translate uniformly to all tasks. For certain bioactivity-related applications, traditional fingerprints remain preferable, as discussed in the Supplementary Materials. More generally, while HDF demonstrates greater consistency and stability across datasets and model types, performance remains dataset-dependent: for some properties, Morgan fingerprints achieve higher accuracy when sufficient dimensionality is available, as HDF performance saturates earlier with increasing embedding size.  

Furthermore, HDF can easily be extended by further atomic or molecular attributes without any training, as we show for stereochemistry in Supplementary Section~\siSecStereo{}. The algebraic structure of hyperdimensional computing also opens directions beyond property prediction. The approximate invertibility of the binding and bundling operations may enable reconstruction of molecular substructures from fingerprint vectors, with potential applications in interpretability and generative molecular design.


\section*{Methods}


\subsection*{Problem statement}

Molecular structures can be naturally represented as undirected graphs \(G = (\mathcal{V}, \mathcal{E})\) consisting of a node set ${\mathcal{V} = \{1, \dots, i, j, \dots, V \}}$ and an edge set ${\mathcal{E} \subseteq \mathcal{V} \times \mathcal{V}}$, where each node corresponds to an atom and each edge to a chemical bond.  Different molecules are thus represented by graphs that vary in order (i.e.\ different numbers of atoms \(|\mathcal{V}|\)), composition and connectivity patterns.  However, many downstream tasks in cheminformatics—such as property prediction, similarity search, and molecular optimization—benefit from using fixed‐length vector inputs \cite{wighReviewMolecularRepresentation2022,davidMolecularRepresentationsAIdriven2020}.

To facilitate computational tasks, especially when the dataset is of limited size, molecular graphs must be transformed into more tractable representations. Specifically, we aim to encode each molecule \(m \in \mathcal{M} \), where $\mathcal{M}$ denotes the space of all valid molecular graphs, into a fixed-dimensional vector $\mathbf{g}$. Formally, this requires defining a mapping
\begin{equation}
    \label{eq:mapping}
    f \;:\; \mathcal{M} \;\longrightarrow\; \mathbb{R}^D,\qquad
    m \;\mapsto\; \mathbf{g},
\end{equation}
where \(\mathbf{g}\in\mathbb{R}^D\) is an embedding that encapsulates the chemical and topological properties of the molecule \(m\). To be broadly applicable, the mapping in Equation~(\ref{eq:mapping}) should
(i) produce an output of fixed dimension \(D\) independent of the number of atoms in \(m\),
(ii) remain invariant under any permutation of the atom indices,
(iii) preserve both local atomic environments (such as atom types and bond orders) and global molecular topology, and
(iv) be computable in time polynomial in the number $|\mathcal{V}|$ of atoms.

An additional desirable property, not exhibited by most of the current methods relying on fixed-length vector representations, is partial invertibility, i.e. the ability to recover structural information about the original molecule from the embedding.

\subsection*{Hyperdimensional computing}
\label{sec:hdc}

Hyperdimensional computing (HDC) \cite{kanervaHyperdimensionalComputingIntroduction2009,kleyko2022survey,heddesHyperdimensionalComputingFramework2024} is a computational paradigm inspired by brain-like mechanisms of information storage. HDC represents concepts as high-dimensional vectors, typically with \(D\gtrsim10^4\), known as hypervectors (HV), which encode information in a distributed and robust manner. Compositional structures are then modeled through a defined set of algebraic operations on the hypervectors, enabling the formation of associative memories.

HDC leverages a mathematical phenomenon observed in high-dimensional vector spaces
\cite{ledoux2001concentration}, where randomly sampled vectors tend to be nearly orthogonal, i.e. $\mathbf{x}\cdot\mathbf{y}\approx 0$. This property lies at the core of the representational power of hypervectors. While the number of exactly orthogonal dimensions in a vector space equals its dimensionality $D$, the number of nearly orthogonal directions grows exponentially with $D$ and their properties increasingly resemble those of perfectly orthogonal vectors. This behavior, often referred to as the blessing of dimensionality \cite{gorban2018blessing}, inherently facilitates the separation of data, enabling hypervectors to serve as highly distinct and robust symbols in high-capacity, noise-tolerant representations.

While multiple different implementations of HDC can be found in the literature, all HDC models are defined by four core components:
\begin{itemize}
    \item a \emph{mapping}  of the fundamental objects—e.g. atoms for molecular representations—to atomic HVs, which may be real-valued, binary, integer-based, or other types, depending on the specific model. Let $\mathcal{H}$ denote this hypervector space;
    \item a \emph{similarity} function $\mathrm{sim}\colon \mathcal{H} \times \mathcal{H} \to \mathbb{R}$, such as cosine similarity, the dot product, or a transformed distance-based kernel (e.g. exponential inverse Euclidean distance);
    \item a \emph{bundling} operation  $\oplus\colon \mathcal{H} \times \mathcal{H} \to \mathcal{H}$, which preserves unstructured similarity—i.e. given two HVs $\mathbf{x}$ and $\mathbf{y}$, the composite output $\mathbf{q} = \mathbf{x} \oplus \mathbf{y}$ remains structurally proximal to its operands under $\mathrm{sim}$, such that $\mathrm{sim}(\mathbf{q}, \mathbf{x})$ and $\mathrm{sim}(\mathbf{q}, \mathbf{y})$ both indicate high similarity;
    \item a \emph{binding} operation $\odot\colon \mathcal{H} \times \mathcal{H} \to \mathcal{H}$, which preserves structured similarity. It produces a representation distant from either operand under $\mathrm{sim}$, but preserves pairwise relationships such that $\mathrm{sim}(\mathbf{x}\odot\mathbf{y},\, \mathbf{x'}\odot\mathbf{y'})$ correlates with the input similarities $\mathrm{sim}(\mathbf{x},\mathbf{x'})$ and $\mathrm{sim}(\mathbf{y},\mathbf{y'})$.
\end{itemize}
Given a compositional hypervector, it is often desirable to recover both the atomic HVs involved in its formation and its compositional structure. This recovery is typically feasible with prior knowledge of the set of potential input HVs and the operations used during the compositional process, although the reconstruction is generally noisy.

Among the various HDC implementations, holographic reduced representations (HRR) \cite{plate1995holographic} offer particularly favorable properties for continuous-valued data: the binding operation is approximately invertible, enabling partial recovery of compositional structure, and similarity computations yield smooth, graded responses rather than discrete outcomes. 

In HRR, the atomic HVs for representing dissimilar objects are real-valued vectors, with their components independently sampled from an isotropic normal distribution with zero mean and variance \(1/D\), where \(D\) is the dimensionality of the vectorial space. For large \(D\), the Euclidean norm of the generated atomic HVs approaches unity. For HDF, we instead use atomic HVs whose Fourier coefficients all have unit magnitude, as in the Fourier variant of HRR (FHRR)~\cite{kleyko2022survey}. We construct such an HV by sampling the phases of its $D$ Fourier coefficients uniformly at random, under the Hermitian symmetry that makes its inverse discrete Fourier transform real-valued. Since binding by circular convolution multiplies the Fourier coefficients of its operands, the repeated binding during message passing would otherwise multiply random coefficient magnitudes and concentrate the information of an embedding on a few Fourier components, whereas these unitary vectors of HRR keep all magnitudes at one under binding~\cite{plate1995holographic}.

The similarity measure between HVs is typically defined as either the dot product or the cosine similarity. The bundling operation is performed as component-wise addition; therefore, normalization is often applied after superposition to preserve the unit norm.

The binding operation in HRR is defined via circular convolution, a mathematical operation that projects the outer product of two vectors onto the same dimensional space:
\[
\mathbf{p} = \mathbf{x} \odot \mathbf{y} \equiv p_j = \sum_{k=0}^{D-1} y_k \, x_{(j-k) \bmod D}.
\]
Circular convolution is a commutative operation that approximately maintains the unit norms of the input HVs.
Finally, the unbinding operation is performed by circular correlation of the composed HV with one of its input HVs, which allows recovering noisy versions of the other input HVs.

\subsection*{Message passing on graphs}

Message passing serves as a mathematical framework for processing graph-structured data within graph neural networks and related architectures \cite{gilmerNeuralMessagePassing2017,khemaniReviewGraphNeural2024}. For an undirected graph $G=(\mathcal{V},\mathcal{E})$ defined by a set of nodes $\mathcal{V}$ and edges $\mathcal{E}$, message-passing algorithms iteratively update node states by propagating information through the network topology.

The process begins by initializing a state vector $h_i^{(0)}$ for each node $i \in \mathcal{V}$. At each subsequent iteration $l$, every node aggregates the hidden states of its immediate structural neighbors $j \in \mathcal{N}(i)$. This mechanism is formalized by a general update function in Equation~(\ref{eq:mp-update}):

\begin{equation}
    \label{eq:mp-update}
    h_i^{(l+1)} = \text{UPDATE} \left( h_i^{(l)}, \text{AGGREGATE} \left( \{ h_j^{(l)} \mid j \in \mathcal{N}(i) \} \right) \right)
\end{equation}

By repeating this procedure for $L$ iterations, the final node state $h_i^{(L)}$ captures structural features within an $L$-hop receptive field centered on node $i$. When a global representation of the entire graph is required, an auxiliary permutation-invariant readout function is deployed to aggregate the hidden representations of all nodes, as given by Equation~(\ref{eq:readout}):

\begin{equation}
    \label{eq:readout}
    g = \text{READOUT} \left( \{ h_i^{(L)} \mid i \in \mathcal{V} \} \right)
\end{equation}


\subsection*{Hyperdimensional graph encoding}

We encode each molecule $m$ as a fixed-length HV, $\mathbf{h} \in \mathbb{R}^D$, by combining principles from HDC and message-passing. The encoding process can be divided into four stages: node attribute embedding, iterative message passing, global attribute embedding, and global aggregation.

\subsection*{Representation of attributes}
To represent molecular properties, we distinguish between categorical and numerical attributes, ensuring that distinct attributes occupy nearly orthogonal subspaces while relevant numerical relationships are preserved. Categorical attributes, such as atom types, are mapped to nearly orthogonal atomic HVs whose Fourier coefficients all have unit magnitude, as described above. Conversely, numerical attributes, such as hydrogen counts, bond degrees, and global graph metrics, are encoded to preserve internal similarity metrics. Specifically, while the HVs of any two distinct attributes remain nearly orthogonal, the encodings within a given numerical attribute space are locality-preserving: values that are numerically close map to HVs with high inner-product similarity, which falls off with the distance between the values according to a sinc kernel~\cite{frady2022computing}.

For a numerical attribute $x \in \mathbb{R}$, \emph{fractional power encoding}~\cite{komer2019neural,frady2022computing} maps it to a hypervector $\mathbf{h}(x)$ using a base hypervector $\mathbf{\Phi}$ and a characteristic bandwidth $\sigma$. This is performed in the Fourier domain to ensure that distances in the Hilbert space correlate with the magnitude of the difference between values, as given by Equation~(\ref{eq:fpe}):
\begin{equation}
    \label{eq:fpe}
    \mathbf{h}(x) = \mathcal{F}^{-1} \left( \mathcal{F}(\mathbf{\Phi})^{x/\sigma} \right)
\end{equation}
where the base hypervector $\mathbf{\Phi}$ is generated with a Hermitian symmetric spectrum to ensure $\mathbf{h}(x)$ remains real-valued and, like the HVs of the categorical attributes, has Fourier coefficients of unit magnitude. The bandwidth $\sigma$ controls the similarity decay: values \(x\) and \(x'\) with \(|x-x'|\ll\sigma\) map to highly similar HVs, while differences much larger than \(\sigma\) yield near-orthogonal representations. Unlike the categorical dictionaries used for atomic features, this construction encodes ordinal proximity and is therefore well suited to real- or integer-valued descriptors whose magnitude carries meaning.

\subsection*{Node attribute embedding}

For each atom $i \in \mathcal{V}$, we consider $K$ local atomic attributes $a_{i,1}, \ldots, a_{i,K}$, each of which is mapped to an attribute-specific HV $\mathbf{h}^{(k)}_{a_{i,k}}$ using either categorical or fractional power encoding. In this work, we use the chemical element $e_i$, the number of bonded hydrogen atoms $h_i$ and the heavy-atom bond degree $b_i$:

\begin{itemize}
    \item Atom type ($\mathbf{h}_{e_i}^{\mathrm{atom}}$): categorical HV representing the chemical element $e_i$;
    \item Hydrogen count ($\mathbf{h}_{h_i}^{\mathrm{hs}}$): fractional power encoding of the hydrogen count $h_i$;
    \item Bond count ($\mathbf{h}_{b_i}^{\mathrm{bonds}}$): fractional power encoding of the heavy-atom bond degree $b_i$.
\end{itemize}

In addition, the Cahn--Ingold--Prelog (CIP) stereo descriptors~\cite{cahnSpecificationMolecularChirality1966,hansonAlgorithmicAnalysisCahnIngoldPrelog2018} of the atoms can optionally be encoded as a further categorical attribute, which separates stereoisomers while the fingerprints of all molecules without specified stereo elements remain unchanged (Supplementary Section~\siSecStereo{}). All results of the main text are obtained without this attribute.

These attribute HVs are combined using the HRR binding operation defined by circular convolution, as given by Equation~(\ref{eq:bind}):
\begin{equation}
    \label{eq:bind}
    \mathbf{u} \odot \mathbf{v}
    =
    \mathcal{F}^{-1}
    \left(
        \mathcal{F}(\mathbf{u})
        \cdot
        \mathcal{F}(\mathbf{v})
    \right),
\end{equation}
where multiplication in the Fourier domain is performed element-wise, $\mathbf{u}, \mathbf{v} \in \mathbb{R}^D$, and $\mathcal{F}$ denotes the Fourier transform operator.

The initial node representation is then constructed as the binding of the HVs of all its atomic attributes,
\begin{equation}
    \label{eq:node-init-general}
    \mathbf{h}_i^{(0)}
    =
    \text{normalize}
    \left(
        \bigodot_{k=1}^{K} \mathbf{h}^{(k)}_{a_{i,k}}
    \right),
\end{equation}
which, for the three attributes used in this work, reads
\begin{equation}
    \label{eq:node-init}
    \mathbf{h}_i^{(0)}
    =
    \text{normalize}
    \left(
        \mathbf{h}_{e_i}^{\mathrm{atom}}
        \odot
        \mathbf{h}_{h_i}^{\mathrm{hs}}
        \odot
        \mathbf{h}_{b_i}^{\mathrm{bonds}}
    \right).
\end{equation}

The initialized node HVs of Equation~(\ref{eq:node-init}) therefore encode atomic identity together with immediate local structural attributes within a unified distributed representation. These node states subsequently serve as the basis for iterative message passing, through which broader topological context is progressively incorporated into the molecular representation.

\subsection*{Hyperdimensional message passing}
To propagate structural information, we perform \(L\) iterations of message passing. At iteration \(l\), each node embedding is updated by
\begin{equation}
\label{eq:mp}
\mathbf{h}_i^{(l+1)}
= \mathrm{normalize}\bigl(\sum_{j\in\mathcal{N}(i)}
\bigl(\mathbf{h}_i^{(l)} \odot \mathbf{h}_j^{(l)}\bigr)\bigr)\,,
\end{equation}
where \(\mathcal{N}(i)\) denotes the set of neighbors of node \(i\), and \(\mathrm{normalize}(\cdot)\) scales its argument to the unit norm. The update rule in Equation~(\ref{eq:mp}) binds the current embedding of \(i\) with that of each neighbor and aggregates the results by summation. Since the sum is commutative, the output is invariant under any permutation of the neighbor indices, while the binding preserves the pairwise relational structure between \(i\) and each \(j \in \mathcal{N}(i)\).

The expressivity of message-passing graph representations is commonly characterized by the one-dimensional Weisfeiler--Leman (1-WL) test~\cite{weisfeilerReductionGraphCanonical1968,morrisWeisfeilerLemanGo2019}, which iteratively refines the color of each node based on its own color and the multiset of colors of its neighbors, and graph isomorphism networks \cite{xu2018powerful} attain the expressivity of 1-WL through injective multiset aggregation. Since binding distributes over addition, Equation~(\ref{eq:mp}) can be written as \(\mathbf{h}_i^{(l+1)} = \mathrm{normalize}\bigl(\mathbf{h}_i^{(l)} \odot \sum_{j\in\mathcal{N}(i)} \mathbf{h}_j^{(l)}\bigr)\), which has the same form as the color refinement of 1-WL. By induction over the iterations, nodes that 1-WL cannot distinguish after \(l\) iterations therefore receive identical HVs, such that HDF is at most as expressive as 1-WL on the encoded graph, with the graph diameter as the only additional information.

Within this bound, binding with a fixed node HV and the summation over multisets of neighbor HVs are injective with probability one, independently of the embedding dimension. For a fixed node HV, binding is a linear map whose eigenvalues are the Fourier coefficients of that HV, and it is therefore invertible unless one of these coefficients vanishes, which is an event of probability zero for randomly sampled atomic HVs. Likewise, the summation maps two different multisets of neighbor HVs to different vectors unless a nontrivial integer combination of these HVs vanishes, which again has probability zero. The concentration of measure in high-dimensional spaces instead determines how well distinct encodings are separated, since the similarity of two sums of quasi-orthogonal HVs reflects the overlap of the underlying multisets up to a crosstalk of the order of \(1/\sqrt{D}\). However, the normalization discards the multiplicity of the aggregated messages, which is compensated by explicitly encoding the bond degree into the initial atomic representations and, at the graph level, by the encoding of the graph size.

After \(L\) rounds of message passing, we aggregate the embeddings at each stage of message passing for each node. Specifically, the embedding of node \(i\) is obtained by summing the embeddings from all iterations \(l = 0\) to \(L\), followed by normalization:
\[
\mathbf{h}_i = \mathrm{normalize}\left(\sum_{l=0}^{L} \mathbf{h}_i^{(l)}\right), \quad \forall i \in \mathcal{V}.
\]

In this way, the node embedding \(\mathbf{h}_i\) captures the molecular environment of the corresponding node, summarizing information from its local neighborhood across \(0\) to \(L\) hops. The binding of each node to the sum of its neighbors also appears in two of the four update variants of the graph vector function architecture (GVFA)~\cite{kahawala2026gvfa}, which additionally superimpose the HV of the node itself and apply a permutation in every iteration.

\subsection*{Global attributes embedding}

Beyond local atomic environments, the fingerprint can include \(M\) global attributes \(c_1(G), \ldots, c_M(G)\) of the molecular graph \(G\), each of which is encoded with categorical or fractional power encoding and which are combined through the bundling operation into a single global-attribute HV:
\begin{equation}
    \label{eq:gattr-general}
    \mathbf{g}_{\mathrm{attr}} \;=\; \mathrm{normalize} \left( \sum_{m=1}^{M} \mathbf{h}^{(m)}_{c_m(G)} \right).
\end{equation}

In this work, we use two global attributes, specifically: the \emph{graph size} \(|\mathcal{V}|\), i.e. the total number of heavy atoms in the graph, and the \emph{graph diameter} \(\mathrm{diam}(G)\), defined as the longest shortest path in \(G\), which we both encode using the fractional power encoding scheme. The corresponding HVs, \(\mathbf{h}_{\mathrm{size}}\) and \(\mathbf{h}_{\mathrm{diam}}\), are combined through the bundling operation into a single global-attribute HV, as given by Equation~(\ref{eq:gattr}):
\begin{equation}
    \label{eq:gattr}
    \mathbf{g}_{\mathrm{attr}} \;=\; \mathrm{normalize} \left( \mathbf{h}_{\mathrm{size}} + \mathbf{h}_{\mathrm{diam}} \right).
\end{equation}

\subsection*{Graph aggregation}

The final molecular fingerprint $\mathbf{g}$ is obtained by integrating the node-level embeddings with the global attributes. We first compute a permutation-invariant structural readout $\mathbf{\mathbf{g}_{\mathrm{struc}}}$ by summing the embeddings across all nodes:
\[
\mathbf{\mathbf{g}_{\mathrm{struc}}} = \text{normalize}\left( \sum_{i \in \mathcal{V}} \mathbf{h}_i \right).
\]
The final fingerprint is the normalized sum of the structural readout and the global attribute vector, as given by Equation~(\ref{eq:final}):
\begin{equation}
    \label{eq:final}
    \mathbf{g} = \text{normalize}\left( \mathbf{\mathbf{g}_{\mathrm{struc}}} + \mathbf{g}_{\mathrm{attr}} \right)
\end{equation}
This resulting vector $\mathbf{g} \in \mathbb{R}^D$ is fixed-dimensional, invariant under atom index permutations and can be seamlessly integrated into downstream tasks such as property prediction and similarity search.


\subsection*{Statistics and reproducibility}

Each dataset was split randomly into training, validation and test sets (80\%/10\%/10\%), and all representations were evaluated on the same splits. Unless stated otherwise, experiments were repeated with 10 independent random seeds, values are reported as mean $\pm$ standard deviation, and box plots show the median and the interquartile range with whiskers extending to 1.5 times the interquartile range; the number of repetitions, campaigns or molecules underlying each figure and table is given in its legend. The sample sizes were determined by the available datasets, and no statistical method was used to predetermine them. Apart from the dataset curation and the exclusions described in the Supplementary Information, no data were excluded from the analyses.

In the dataset comparisons, the error of each representation was divided by the lowest error for the respective property, and these error ratios were summarized across properties by their geometric mean~\cite{flemingHowNotLie1986}, given as a relative deviation from the best result, and by the number of properties with a deviation of at most 10\%~\cite{dolanBenchmarkingOptimizationSoftware2002}. Representations were compared across properties with a Friedman test, followed by two-sided Wilcoxon signed-rank tests between HDF and each baseline on the logarithms of the per-property error ratios, with Holm correction~\cite{demsarStatisticalComparisonsClassifiers2006}. In the Bayesian-optimization benchmark, HDF and Morgan fingerprints were compared at each embedding size with two-sided Wilcoxon signed-rank tests on the paired campaigns, Holm-corrected within each objective.

\subsection*{Use of large language models}
During the preparation of this work, the authors used Claude (Anthropic) to assist in drafting the manuscript, to copy-edit and improve the readability of the text, and to write parts of the code used in this study, including scripts for data visualization. Claude was not used to analyse or interpret the scientific data, and no figures or images were created using generative AI; all figures were generated programmatically from the study's results. The authors reviewed, verified, and edited all AI-generated text and code, and take full responsibility for the content of this work. Large language models are not a component of the hyperdimensional fingerprint method presented in this study.


\backmatter

\bmhead{Data availability}
\cfgDataAvailability

\bmhead{Code availability}
\cfgCodeAvailability

\bibliography{references}

\bmhead{Acknowledgements}
\cfgAcknowledgements

\bmhead{Author contributions}
\cfgAuthorContributions

\bmhead{Competing interests}
\cfgCompetingInterests


\end{document}